\documentclass[letterpaper]{article} 
\usepackage{aaai2027}
\usepackage[hyphens]{url}  
\usepackage{graphicx} 
\usepackage{natbib}  
\usepackage{caption} 
\usepackage[table]{xcolor}
\definecolor{oursrow}{RGB}{245,237,226} 
\usepackage{algorithm}
\usepackage{graphicx}
\usepackage{amsmath}
\usepackage{amsfonts}
\usepackage{algorithm}
\usepackage{algpseudocode}
\usepackage{newfloat}
\usepackage{listings}
\DeclareCaptionStyle{ruled}{labelfont=normalfont,labelsep=colon,strut=off} 
\floatstyle{ruled}
\newfloat{listing}{tb}{lst}{}
\floatname{listing}{Listing}

\usepackage{booktabs}

\title{Boosting Deepresearch and LongContext Ability with Self-Generated Deepresearch Rollouts Traces}

\author{
    Zihan Wang\textsuperscript{\rm 1}\equalcontrib,
    Hao Wang\textsuperscript{\rm 2}\equalcontrib,
    Boyuan Jiang\textsuperscript{\rm 3},
    Yiqun Zhang\textsuperscript{\rm 1},
    Shi Feng\textsuperscript{\rm 1}\corresponding,\\
    Xiaocui Yang\textsuperscript{\rm 1},
    Yiwen Ye\textsuperscript{\rm 4},
    Jianghang Lin\textsuperscript{\rm 5},
    Xiaozhong Ji\textsuperscript{\rm 6},
    Jinghao Lin\textsuperscript{\rm 1},
    Kai Wu\textsuperscript{\rm 7}
}

\affiliations{
    \textsuperscript{\rm 1}Northeastern University, 
    \textsuperscript{\rm 2}City University of Hong Kong,
    \textsuperscript{\rm 3}Zhejiang University, \\
    \textsuperscript{\rm 4}Northwestern Polytechnical University, 
    \textsuperscript{\rm 5}Xiamen University, 
    \textsuperscript{\rm 6}Nanjing University, 
    \textsuperscript{\rm 7}Tongji University
}

\begin{document}

\maketitle

\begin{abstract}
Deepresearch (DR) agents interact with real-world web environments through multi-turn search and visit, causing their contexts to grow rapidly over time. 
We observe that, even after DR Agentic Reinforcement Learning (DR-RL), 61.6\% of the model's remaining prediction errors can still be attributed to insufficient long-context understanding, including long-context hallucination and failures in cross-document evidence integration. 
It motivates us to further break the bottleneck of DR-RL by strengthening the model's long-context ability. 
However, effective LongContext training requires more than simply increasing context length. 
To bridge the data gap, we propose `DR Rollouts to LongContext-QA (\textbf{DR-to-Long})'. 
The method repurposes DR-RL trajectories, which naturally contain search histories, visited webpages, evidence snippets, and final-answer supervision. 
It then replaces the compact snippets and webpage summaries in each trajectory with the full contents of their corresponding URLs, producing substantially longer multi-document contexts while preserving the original evidence relationships. 
Building on \textbf{DR-to-Long}, we introduce \textbf{DLD (DR $\rightarrow$ LongQA $\rightarrow$ DR)-RL}. 
\textbf{DLD-RL} first performs a short DR-RL stage to collect rollout trajectories, which are then converted into LongQA instances at zero annotation cost. 
The model is subsequently optimized with LongQA-RL to strengthen LongContext ability, followed by full DR-RL to continue improving its DR capability. 
Experiments show that DLD-RL outperforms standard DR-RL by 7.3\% on three Deepresearch benchmarks and improves performance by 13.5\% on three long-context benchmarks. 

\end{abstract}


\section{Introduction}
\begin{figure}[t]
    \centering
    \resizebox{\columnwidth}{!}{
        \includegraphics{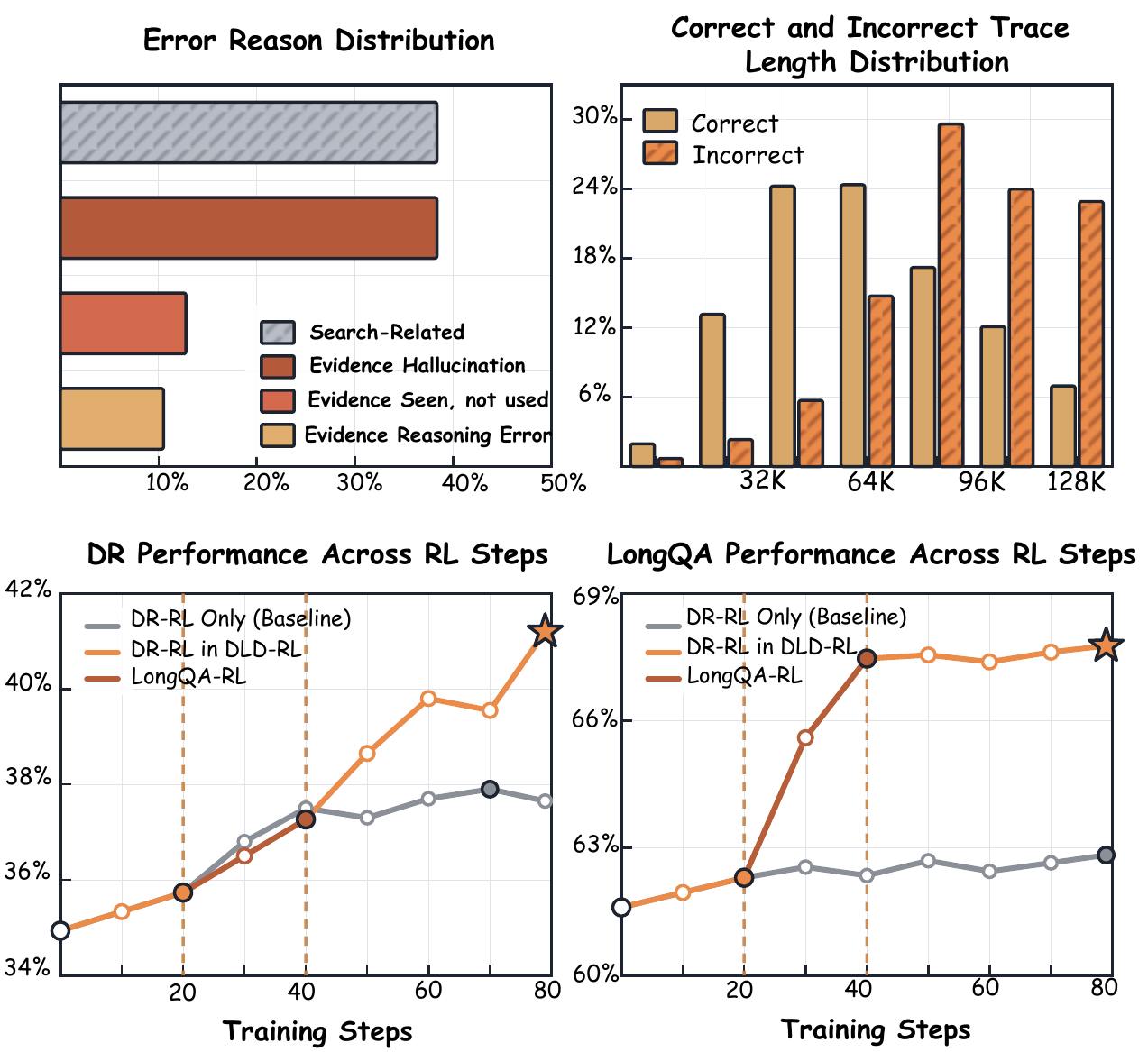}
    }
    \caption{Top left: Analysis of incorrect reasons in DR tasks. Top right: Distribution of trace lengths for correct and incorrect trajectories during training. Bottom: Performance comparison between DLD-RL and DR-RL only.}
    \label{fig:intro}
\end{figure}

Deepresearch (DR) has attracted increasing attention as a representative agentic task~\cite{team2026kimi,zeng2025glm,xu2025comprehensive,xu2026deepseek}. To answer complex questions, models must repeatedly search and browse the Web, gather evidence from multiple sources, and synthesize a final response~\cite{li2026webthinker,wu2026webdancer,li2025websailor,lu2025deepdive}. 
Such interactions can easily exceed 100 steps~\cite{team2026mirothinker,team2025mirothinker,team2025tongyi}, causing retrieved evidence and intermediate reasoning to accumulate continuously in the context. 
At latter steps, the model must reorganize this growing context to determine its next tool action~\cite{yao2022react}. 
This suggests that LongContext understanding is an essential component of DR capability. 
Specifically, we use a stronger language model to analyze the 1,000 trajectories of erroneous predictions made by several existing DR models on the evaluated benchmarks~\cite{team2025tongyi,team2026dr}. 
The analysis shows that 61.6\% of these errors are associated with insufficie nt long-context understanding. 
Typical failures include generating claims unsupported by the retrieved evidence and failing to connect evidence across documents even when the information required for the answer is already present in the context, as illustrated in the upper-left panel of Fig.~\ref{fig:intro}.
Moreover, recent context-management methods further support this view: by alleviating the burden of LongContext processing, they can improve DR performance~\cite{wu2025resum,sun2025scaling,lu2026longseeker}. 

Researchers typically rely on DR Agentic Reinforcement Learning (DR-RL) to push a model's Deepresearch capability toward its performance ceiling~\cite{lu2025deepdive,chu2026redsearcher,team2025tongyi}. 
However, we observe that such improvements in DR performance lead to only marginal gains in long-context understanding. 
We hypothesize that positive training signals in DR-RL predominantly come from shorter trajectories, where accurate early decisions allow the model to quickly retrieve useful evidence and complete the task. 
In contrast, longer trajectories often involve repeated searches, ineffective evidence gathering, or information confusion, and are therefore more likely to fail, as illustrated in the upper-right panel of Fig.~\ref{fig:intro}. Consequently, standard DR-RL receives limited positive supervision from long trajectories and may not sufficiently develop stable decision-making over long contexts. This motivates us to explicitly strengthen long-context understanding to further break the performance bottleneck of DR-RL. 

However, high-quality long-context training data is difficult to construct. Such data must include not only multiple source documents highly relevant to the question, but also challenging distractor documents, implicit cross-document evidence chains, and verifiable final answers~\cite{bai2025longbench,bai2024longbench}. 
These elements are essential for training models to locate and connect evidence across long contexts~\cite{bai2024longalign,wang2024leave,liu2024lost}. 
We realize that DR agents themselves are natural producers of such data. When solving a complex question, a DR agent actively searches for information, visits multiple relevant webpages, filters evidence, and generates a final answer. 
This process naturally provides the key components of long-context training instances: visited webpages serve as evidence documents, retrieved but unused pages provide related yet challenging distractors, and the final answer together with task feedback supplies the training signal. 
Motivated by this observation, we introduce \textbf{DR Rollouts to Long-Context QA (DR-to-Long)}, which transforms existing DR interaction trajectories into multi-document long-context QA (LongQA) data with no additional annotation cost. 
We pair the original question with the expanded webpage context to construct a long-context QA instance. 
For successful rollouts, the correct final answer provides evidence that the retrieved context already contains sufficient information to solve the question, allowing these traces to be converted directly. 
Failed rollouts, however, are potentially more valuable because they often contain noisier evidence and more challenging distractors, but they may lack some information required for deriving the correct answer. We therefore augment the expanded context of a failed rollout with the compact, high-value search snippets and webpage summaries from a successful rollout of the same question. It saves the difficulty of the failed trajectory while ensuring answer sufficiency. 
Our experiments show that such augmented failed-rollout instances are more challenging and yield better training effectiveness than data constructed solely from successful rollouts. 
By controlling the number of webpages expanded, our method can flexibly generate contexts from 64K--128K tokens in training and scale to million-token instances without any annotation. 

\textbf{DR-to-Long} naturally leads to our training framework, \textbf{DLD-RL (DR $\rightarrow$ LongQA $\rightarrow$ DR)}. 
DLD-RL first performs a short stage of DR-RL to generate rollout traces, which are subsequently converted into LongContext-QA data by \textbf{DR-to-Long}. 
The model is then trained on these instances with longQA-RL to strengthen its long-context understanding ability, after which full DR-RL is resumed to further improve its agentic performance. 
Compared with continuously applying Deepresearch RL only, DLD-RL achieves consistently better performance and surpasses the original DR-RL performance ceiling. 
Meanwhile, the intermediate LongQA-RL stage substantially improves the model's long-context reasoning capability. 

We evaluate DLD-RL on two models initialized from Deepresearch-SFT checkpoints. 
Across three Deepresearch benchmarks, DLD-RL achieves an average relative improvement of approximately 7.3\% over standard Deepresearch RL. It also delivers an average relative improvement of approximately 13.5\% across three long-context benchmarks. 
Fig. ~\ref{fig:intro} shows the score fluctuations of pure DR-RL and DLD-RL on two types of benchmarks. 
Further analysis shows that LongQA-RL does more than improve long-context benchmark scores: it also meaningfully changes the behavior of the DR agent. For example, the resulting model invokes the \textsc{Visit} tool more frequently to inspect the full contents of webpages rather than relying primarily on compact search snippets. These findings reveal a mutually reinforcing relationship between LongContext and DR capability. 
Our contributions are summarized as follows: 
\begin{itemize}
\item We provide empirical evidence that long-context understanding is a major bottleneck for current Deepresearch(DR) agents. 
Our analysis further shows that standard DR reinforcement learning (DR-RL) alone struggles to overcome this limitation. 

\item We propose \textbf{DR-to-Long}, a data-construction method that recompiles rollout trajectories generated during DR-RL into directly usable, multi-document long-context QA instances with nearly no additional annotation cost. It can even produce training instances exceeding 1M tokens. 

\item We introduce \textbf{DLD-RL}, a \textbf{DR $\rightarrow$ LongQA $\rightarrow$ DR} training framework that uses rollouts collected during an initial DR-RL stage to construct long-context training data, interleaves a dedicated LongQA-RL stage, and subsequently resumes full DR-RL. DLD-RL achieves substantial improvements across three Deepresearch benchmarks and three long-context benchmarks. 
\end{itemize}

\section{Related Work}
\paragraph{Deepresearch Agent}
Large language model agents increasingly augment internal knowledge with external information, as exemplified by RAG~\cite{lewis2020retrieval}. Deepresearch extends this paradigm by enabling models to actively search and reason over retrieved evidence~\cite{lu2025deepdive,wang2026deepmed}, often across hundreds of interaction steps~\cite{zeng2025glm}. Benchmarks such as BrowseComp~\cite{wei2025browsecomp} and XBench~\cite{chen2025xbench} evaluate these capabilities in realistic Web environments, requiring extensive tool use, evidence localization, and multi-source reasoning.
Deepresearch trajectory data can be used for Agentic SFT to familiarize models with tool use and multi-turn interaction~\cite{du2026openseeker1,du2026openseeker2}, while Deepresearch RL further improves autonomous search, browsing, and evidence synthesis~\cite{chu2026redsearcher,team2025mirothinker}. However, even after Deepresearch RL, models may still fail due to long-context hallucination, evidence omission, or weak cross-document integration. DLD-RL targets this remaining bottleneck and thereby further improves Deepresearch performance.

\paragraph{LongContext Ability in Long-Horizon Tasks}
As models are expected to solve increasingly complex tasks, they must often decompose them into multiple subproblems and operate over longer time horizons, motivating the growing interest in long-horizon intelligence~\cite{erdogan2025plan,zeng2026glm,wang2025odysseybench,patwardhan2025gdpval}. Modern agent systems can easily involve hundreds or even thousands of interaction steps, causing their accumulated reasoning and observations to rapidly expand the required context~\cite{team2026mirothinker,lee2026meta,desai2026swe}. Stable performance in such settings therefore depends on strong long-context capabilities~\cite{hu2025hiagent,dou2026cllife,dou2026cl}, as evaluated by benchmarks such as LongBench-v2~\cite{bai2025longbench} and CL-Bench~\cite{dou2026cl}.
DeepResearch is a representative long-horizon task requiring multi-document reasoning and evidence synthesis. Its rollouts provide long-context training data without extra annotation, improving both long-context understanding and task performance, with potential transfer to other long-horizon tasks.

\section{Preliminaries}

\paragraph{Problem Settings}
We consider two task settings: Deepresearch-QA (DR QA) and LongContext-QA (Long QA).
In a DR-QA task, the system input consists of the system prompt $P_S$ and the tool specifications $P_T$, denoted as $P_S + P_T$, while the user provides only the question $P_Q$. The tool specifications describe the set of tools available to the model and provide the necessary instructions for using them, including their functions, expected input formats, and returned outputs. 
Based on these specifications, the model can determine which tool to invoke and how to construct a valid tool call during the research process. 
At each interaction step, the model generates a reasoning trace followed by a tool call, denoted as $Res_{\mathrm{think}} + Res_{\mathrm{tool}}$. 
The backend then executes the tool call, returns the resulting observation to the model, and initiates the next interaction round. 
This iterative process continues until the model stops invoking tools and produces a final textual answer like $Res_{\mathrm{think}} + Res_{\mathrm{text}}$ or until the maximum interaction or context budget is reached. 
In a Long QA task, the system input contains the system prompt $P_S$, while the user input consists of the question $P_Q$ and its associated context $P_{\mathrm{Ctx}}$. 
Without interacting with external tools, the model directly generates a reasoning trace followed by a textual answer, denoted as $Res_{\mathrm{think}} + Res_{\mathrm{text}}$.

\paragraph{Tool Settings}
We equip the agent with two commonly used Deepresearch tools: \textsc{Search} and \textsc{Visit}. \textsc{Search} is implemented through Serper\footnote{\url{https://serper.dev/}} and performs keyword-based Web retrieval, returning the snippets and URLs of the top-10 relevant webpages. \textsc{Visit} accesses the webpage associated with a given URL through Jina\footnote{\url{https://jina.ai/}} and returns content summarized according to a specified summary goal. A separate summarization model is used to condense lengthy webpages. 
Consequently, the agent only observes compressed representations of the underlying webpages, whether through the snippets returned by \textsc{Search} or the summaries produced by \textsc{Visit}. 
\textbf{DR-to-Long} expands these traces into long-context data by replacing such compressed observations with the full contents of their corresponding webpages. 

\section{DLD (DR -> Long -> DR)-RL}
\begin{figure*}[t]
    \centering
    \resizebox{\textwidth}{!}{
        \includegraphics{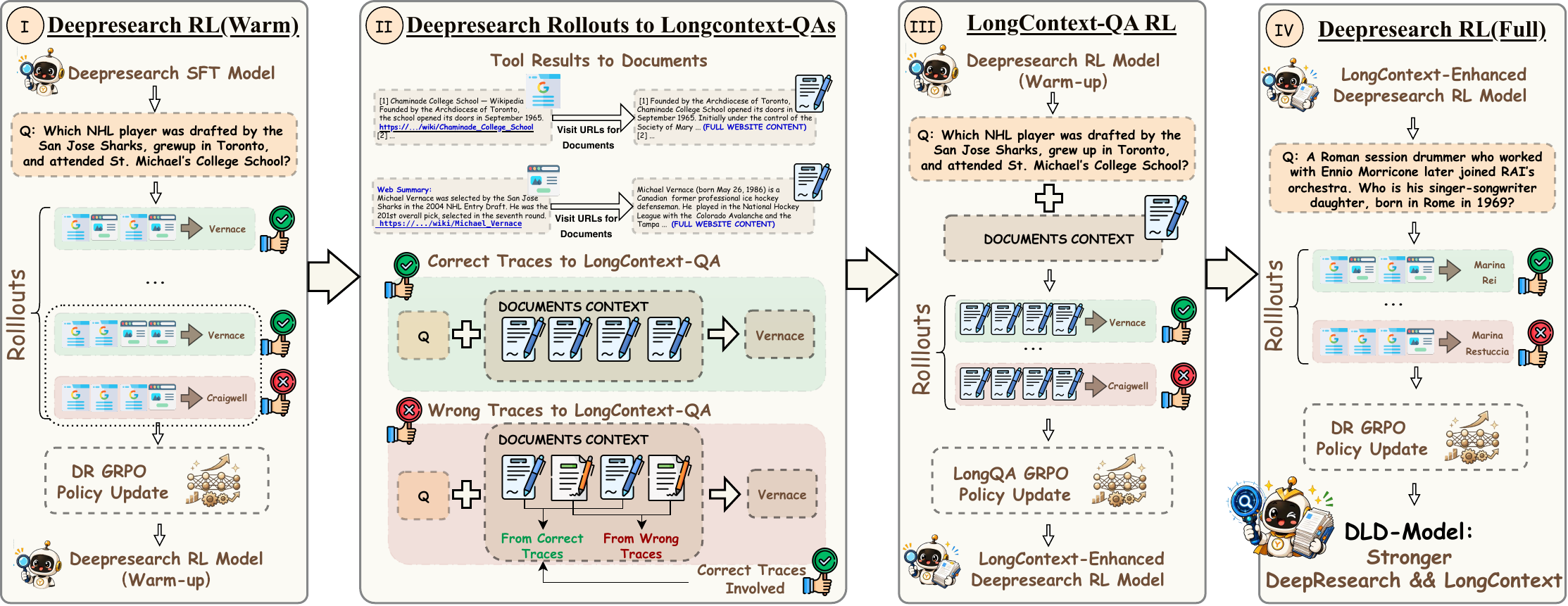}
    }
    \caption{Overview of the proposed DLD-RL framework.}
    \label{fig:dld-rl}
\end{figure*}

As illustrated in Fig.~\ref{fig:dld-rl} and Alg.~\ref{alg:lrrl}, DLD-RL consists of four stages. First, starting from a DeepResearch-SFT model, we perform a short stage of standard GRPO-based DeepResearch RL to collect rollout trajectories. Second, these trajectories are converted into LongContext-QA training data. Third, we continue training from the resulting warm-up DeepResearch policy and conduct LongContext-QA RL to strengthen its long-context capabilities. Finally, the long-context-enhanced model resumes DeepResearch RL and is further optimized until performance saturates. 

\begin{algorithm}[t]
\caption{DLD-RL}
\label{alg:lrrl}
\small
\begin{algorithmic}
\Require Agentic-SFT policy $\pi_{\mathrm{SFT}}$, DR dataset
$\mathcal{D}_{\mathrm{DR}}$, context limit $L_{\max}$
\Ensure Final policy $\pi^{*}$

\Statex \textbf{Stage 1: Collect DR rollouts}
\State $(\pi_{\mathrm{warm}},\mathcal{R})
\gets \Call{DeepresearchRL}
{\pi_{\mathrm{SFT}},\mathcal{D}_{\mathrm{DR}},
\beta_{\mathrm{KL}}=0}$

\Statex \textbf{Stage 2: Construct LongQA data}
\State $\mathcal{D}_{\mathrm{LongQA}}\gets\emptyset$

\ForAll{$(q,a^*)\in\mathcal{D}_{\mathrm{DR}}$}
    \State $\mathcal{R}_q^{+},\mathcal{R}_q^{-}
    \gets \Call{SplitByReward}{\mathcal{R}_q}$
    \State $E_q^{+}
    \gets \Call{ExtractEvidence}{\mathcal{R}_q^{+}}$

    \ForAll{$\tau^{+}\in\mathcal{R}_q^{+}$}
        \State $C^{+}
        \gets \Call{ExpandURLs}{\tau^{+},L_{\max}}$
        \State $\mathcal{D}_{\mathrm{LongQA}}
        \gets \mathcal{D}_{\mathrm{LongQA}}
        \cup \{(q,C^{+},a^*)\}$
    \EndFor

    \ForAll{$\tau^{-}\in\mathcal{R}_q^{-}$}
        \State $C^{-}
        \gets \Call{ExpandURLs}{\tau^{-},L_{\max}}$
        \State $\widetilde{C}^{-}
        \gets \Call{AugmentContext}
        {C^{-},E_q^{+},L_{\max}}$
        \State $\mathcal{D}_{\mathrm{LongQA}}
        \gets \mathcal{D}_{\mathrm{LongQA}}
        \cup \{(q,\widetilde{C}^{-},a^*)\}$
    \EndFor
\EndFor

\Statex \textbf{Stage 3: Long-context RL with KL regularization}
\State $\pi_{\mathrm{ref}}
\gets \Call{Freeze}{\pi_{\mathrm{warm}}}$
\State $\pi_{\mathrm{L}}
\gets \Call{LongQARL}
{\pi_{\mathrm{warm}},\pi_{\mathrm{ref}},
\mathcal{D}_{\mathrm{LongQA}},
\beta_{\mathrm{KL}}>0}$

\Statex \textbf{Stage 4: Resume full DR-RL without KL}
\State $\pi^{*}
\gets \Call{DeepresearchRL}
{\pi_{\mathrm{L}},\mathcal{D}_{\mathrm{DR}},
\beta_{\mathrm{KL}}=0}$

\State \Return $\pi^{*}$
\end{algorithmic}
\end{algorithm}

\subsection{Deepresearch RL (Warm)}
\label{subsec:dr_rl_warm}
We first perform a short stage of Deepresearch reinforcement learning as a warm-up. Through multi-turn interactions with Web tools, the model learns to search, visit webpages, terminate tool use, and produce a final answer. 
Rather than training the model to convergence at this stage, our primary objective is to collect informative Deepresearch trajectories for the subsequent \textbf{DR-To-Long} conversion. These trajectories provide the search queries, visited URLs, compressed webpage observations, and final responses required to construct long-context question-answering data. Both successful and unsuccessful trajectories are retained, as they expose diverse retrieval paths and webpage evidence that can be reused in the following stage.

We adopt GRPO~\cite{shao2024deepseekmath} to optimize the Deepresearch policy. For each question $q$, we sample a group of trajectories
$\mathcal{T}=\left\{{\tau_i}\right\}_{i=1}^{G}$
from the old policy $\pi_{\theta_{\mathrm{old}}}$. The optimization objective is defined in Eq.~\ref{eq:agentic_grpo}:
\begin{equation}
\mathcal{J}_{\mathrm{DR}}(\theta)=
\mathbb{E}_{q,\mathcal{T}}
\left[
\frac{1}{G}\sum_{i=1}^{G}
\frac{1}{|\tau_i|}
\sum_{t=1}^{|\tau_i|}
\ell(\rho_{i,t},\hat{A}_i)
\right],
\label{eq:agentic_grpo}
\end{equation}
where $\rho_{i,t}$ denotes the probability ratio between the current and old policies,
\begin{equation}
\hat{A}_i =
\frac{
r(\tau_i)-\operatorname{mean}\left({r(\tau_j)}_{j=1}^{G}\right)
}{
\operatorname{std}\left({r(\tau_j)}_{j=1}^{G}\right)+\delta
},
\end{equation}
is the group-normalized advantage, and
$\ell(\rho,A)=\min\left(\rho A,\operatorname{clip}(\rho,1-\epsilon,1+\epsilon)A\right)$ is the clip mechanism. 

We use an LLM-judged binary reward $r$: a trajectory receives a reward of $1$ if its final answer matches the reference answer and $0$ otherwise. 
Traces that exceed the maximum length or fail to produce a valid answer also receive $0$ reward. 
Following common practice in DR RL~\cite{lu2025deepdive,team2025mirothinker}, we omit the KL regularization term.

\subsection{Deepresearch Rollouts to LongContext-QAs}

During Deepresearch GRPO, we sample multiple rollout trajectories for each question. 
Different trajectories often retrieve different webpages, providing diverse sources for constructing long-context training data. 
For a question $q$, the collected rollout trajectories are represented as Eq.~\ref{eq:rollout_set}:
\begin{equation}
\mathcal{T}(q)=\{\tau_i\}_{i=1}^{N},
\label{eq:rollout_set}
\end{equation}
where each trajectory $\tau_i$ consists of interleaved reasoning traces, tool calls, and tool observations. 
For each trajectory, we extract the webpages collected through \textsc{Search} and \textsc{Visit}. 
The retrieved documents are denoted as $D$ as shown in Eq.~\ref{eq:document_collection}:
\begin{equation}
D=\{d_{1},d_{2},...,d_{n}\},
\label{eq:document_collection}
\end{equation}
where each $d_i$ represents an individual retrieved document, including search snippets, webpage summaries, or reconstructed webpage contents. 
We concatenate these documents to construct a long context $C_i$, as defined in Eq.~\ref{eq:context_construction}:
\begin{equation}
C=[d_1;d_2;...;d_n],
\label{eq:context_construction}
\end{equation}
which is then paired with the original question and answer to form a LongContext-QA instance $(q,C,a^*)$, which is exactly the training sample, where $q$ is the original question, $C$ is the constructed long context, and $a^{*}$ is the original answer.

For successful trajectories, $C_i$ directly represents the evidence collected by the agent to solve the original Deepresearch task. For unsuccessful trajectories, we construct contexts in the same manner. Since these contexts correspond to cases where the model failed, they provide challenging examples involving incomplete evidence selection or cross-document reasoning failures. We further combine unsuccessful contexts with search/visit results in successful contexts from the same question to create harder training instances. 

\begin{figure}[h]
    \centering
    \resizebox{\columnwidth}{!}{
    \includegraphics{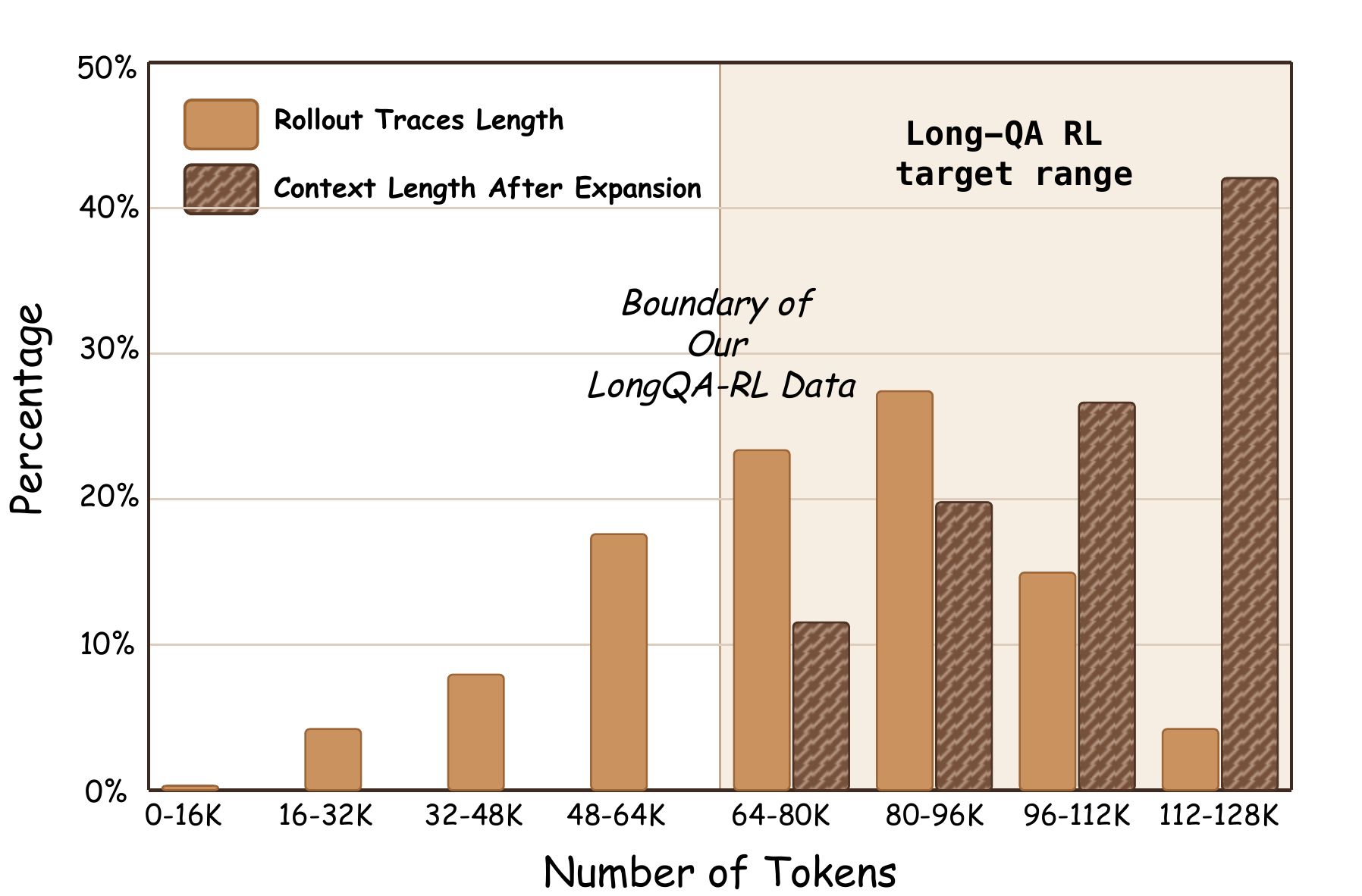}
    }
    \caption{Length distributions of the original Deepresearch rollout trajectories and the LongContext-QA instances constructed through DR-To-Long. We control the LongContext-QA's context length to 64k-128k.}
    \label{fig:context_distribution}
\end{figure}

Finally, we revisit the URLs contained in each trajectory and retrieve the original webpage content, replacing snippets or summaries to further expand the context. 
The length comparison between the original rollout traces and the resulting LongContext-QA data is shown in Fig.~\ref{fig:context_distribution}. 

\subsection{LongContext-QA RL}
After the DR-To-Long conversion, the resulting LongContext-QA instances are used to further optimize the model. 
Unlike Deepresearch RL, no \textsc{Search} or \textsc{Visit} tools are available at this stage; instead, the model directly generates an answer based on the provided long context. 
GRPO~\cite{shao2024deepseekmath} is applied to the constructed training data. 
Rather than optimizing tool-use behaviors, this stage focuses on improving long-context comprehension, evidence localization, and cross-document integration. 
The reward is binary, too. 
And invalid or truncated responses receive zero reward.
To retain the tool-use and multi-turn capabilities learned during DeepResearch RL (Warm), we keep a frozen reference policy $\pi_{\mathrm{ref}}$ and add KL regularization to the objective in Eq.~\ref{eq:longqa_grpo}.
\begin{equation}
\begin{split}
\mathcal{J}_{\mathrm{Long}}(\theta)=
\mathbb{E}_{x,\mathcal{Y}}
\Bigg[
&\frac{1}{G}
\sum_{i=1}^{G}
\frac{1}{|y_i|}
\sum_{t=1}^{|y_i|}
\Big(
\ell(\rho_{i,t},\hat{A}_i) \\
&\quad -
\beta
D_{\mathrm{KL}}
\left(
\pi_{\theta}
\Vert
\pi_{\mathrm{ref}}
\right)_{i,t}
\Big)
\Bigg],
\end{split}
\label{eq:longqa_grpo}
\end{equation} 
where $D_{\mathrm{KL}}$ represents the low-variance KL estimator, and $\beta$ controls the KL regularization strength. Both the initial policy and reference policy are initialized from the Warm-up Deepresearch RL checkpoint, while $\pi_{\mathrm{ref}}$ remains frozen during LongContext-QA RL. 

\subsection{Deepresearch RL (Full)}
Through the preceding stages, DLD-RL first strengthens the model's Deepresearch capability and then substantially improves its long-context QA performance. However, because LongQA-RL primarily optimizes reasoning over long inputs, the resulting model may not yet achieve its full potential on Deepresearch tasks. We therefore use this model to initialize the final Deepresearch RL stage, which serves as an annealing phase to realign the policy with the Deepresearch objective. 
Specifically, we resume training with the same optimization procedure described in Sec.~4.2 and continue until performance saturates. Together, the four stages of DLD-RL form a closed training loop without additional annotation. 
\section{Experiments}
\subsection{Benchmarks and Baselines}
\paragraph{Benchmarks}
We evaluate our models and several baselines on both Deepresearch(DR) and LongContext question-answering tasks. 
For Deepresearch evaluation, we use three representative benchmarks: BrowseComp-300~\cite{wei2025browsecomp}, SEAL-0~\cite{pham2025sealqa}, and the more challenging XBench-2510~\cite{chen2025xbench}. 
For long-context evaluation, we use LongBench-v2~\cite{bai2025longbench}, Frames~\cite{krishna2025fact}, and LongReason~\cite{ling2025longreason}. 
Together, these benchmarks assess Web search and evidence synthesis as well as long-context comprehension and cross-document reasoning.
\paragraph{Baselines}
For DR evaluation, we compare against several similarly sized DR models, including DeepDive~\cite{lu2025deepdive}, WebSailor~\cite{li2025websailor}, OffSeeker~\cite{zhou2026offseeker}, WebExplorer~\cite{liu2025webexplorer}, and AgentCPM-Explore~\cite{chen2026agentcpm}. For long-context evaluation, we report the performance of both general-purpose base models and Deepresearch models. The base-model baselines include Qwen3-4B/30B-Thinking-2507~\cite{yang2025qwen3} and Qwen3.5-9B-Instruct~\cite{qwen3.5}. 

\subsection{Detailed Settings}
\paragraph{Backbones}
We conduct subsequent reinforcement learning on two backbone models: DR-Venus-4B-SFT~\cite{team2026dr}, an existing model that has undergone Deepresearch supervised fine-tuning, and another model that we fine-tune for Deepresearch using the REDSearcher-SFT-10K~\cite{chu2026redsearcher} dataset. 

\paragraph{Training Parameters}
During reinforcement learning, each batch consists of eight queries, with eight sampled responses generated for each query. 
For LongQA-RL, we set the KL-divergence coefficient to 0.005. The training queries are drawn from the REDSearcher~\cite{chu2026redsearcher} dataset. 
For a fair comparison, DLD-RL and the DR-RL-only baseline use the same total number of RL optimization steps. In DLD-RL, this budget is distributed across its three training stages ($DR \rightarrow Long \rightarrow DR$), while the baseline uses it entirely for continuous DR-RL. 
DeepSeek-V4-Flash~\cite{xu2026deepseek} is used to judge whether the model’s final answer matches the reference answer and give the reward; responses without a valid final answer are marked as incorrect. 

\paragraph{Test Parameters}
We evaluate each benchmark three times and report the average score. For DeepResearch evaluation, we set the maximum number of interaction turns to 150, the maximum context length to 128K tokens, and the maximum output length per turn to 9.6K tokens. For long-context evaluation, the maximum input and output lengths are set to 128K and 32K tokens, respectively. For instances whose input exceeds the 128K-token limit, we follow the commonly adopted middle-truncation strategy~\cite{wang2025cstree,bai2025longbench,soh2025you,liu2025reattention}, retaining the first 64K and the last 64K tokens of the original context. 
During evaluation, we adopt the same judging protocol as that used to compute the training reward. 

\begin{table}[t]
\centering
\resizebox{\columnwidth}{!}{%
\begin{tabular}{@{}lccc@{}}
\toprule
\textbf{Model}
& \shortstack{\textbf{BrowseComp}}
& \shortstack{\textbf{XBench-2510}}
& \shortstack{\textbf{Seal-0}} \\
\midrule

DeepDive-9B-SFT
& 5.6 & -- & 15.2 \\

DeepDive-9B-RL
& 6.3 & -- & 12.2 \\

WebSailor-7B
& 6.7 & -- & -- \\

OffSeeker-8B-SFT
& 10.6 & -- & -- \\

OffSeeker-8B-DPO
& 12.8 & -- & -- \\

WebExplorer-8B-RL
& 15.7 & 23.0 & -- \\

AgentCPM-Explore-4B
& 24.1 & 32.0 & 33.3 \\

\midrule

DR-Venus-4B-SFT
  & $26.8\,{\scriptstyle\pm 2.1}$
  & $31.0\,{\scriptstyle\pm 3.5}$
  & $31.8\,{\scriptstyle\pm 3.6}$ \\

  \quad $+$ DR-RL Only
  & $\underline{29.7}\,{\scriptstyle\pm 3.4}$
  & $\underline{33.0}\,{\scriptstyle\pm 2.0}$
  & $\underline{38.7}\,{\scriptstyle\pm 3.9}$ \\

  \quad $+$ \textbf{(Ours) D(Warm)-RL}
  & $27.3\,{\scriptstyle\pm 2.6}$
  & $29.7\,{\scriptstyle\pm 3.8}$
  & $33.6\,{\scriptstyle\pm 2.8}$ \\

  \quad $+$ \textbf{(Ours) D(Warm)L-RL}
  & $27.5\,{\scriptstyle\pm 3.2}$
  & $\underline{33.3}\,{\scriptstyle\pm 2.5}$
  & $36.9\,{\scriptstyle\pm 3.6}$ \\

  \quad $+$ \textbf{(Ours) DLD-RL}
  & $\mathbf{31.7}\,{\scriptstyle\pm 2.8}$
  & $\mathbf{35.0}\,{\scriptstyle\pm 4.0}$
  & $\mathbf{39.6}\,{\scriptstyle\pm 3.1}$ \\

  \midrule

  Qwen3.5-9B-Instruct-DR-SFT
  & $31.4\,{\scriptstyle\pm 3.1}$
  & $32.0\,{\scriptstyle\pm 1.7}$
  & $41.4\,{\scriptstyle\pm 2.4}$ \\

  \quad $+$ DR-RL Only
  & $\underline{34.6}\,{\scriptstyle\pm 2.4}$
  & $\underline{35.0}\,{\scriptstyle\pm 3.6}$
  & $44.1\,{\scriptstyle\pm 3.2}$ \\

  \quad $+$ \textbf{(Ours) D(Warm)-RL}
  & $32.1\,{\scriptstyle\pm 3.6}$
  & $32.3\,{\scriptstyle\pm 2.9}$
  & $43.2\,{\scriptstyle\pm 3.9}$ \\

  \quad $+$ \textbf{(Ours) D(Warm)L-RL}
  & $31.9\,{\scriptstyle\pm 2.9}$
  & $34.3\,{\scriptstyle\pm 3.2}$
  & $\underline{45.9}\,{\scriptstyle\pm 2.7}$ \\

  \quad $+$ \textbf{(Ours) DLD-RL}
  & $\mathbf{36.8}\,{\scriptstyle\pm 3.8}$
  & $\mathbf{40.0}\,{\scriptstyle\pm 3.0}$
  & $\mathbf{47.7}\,{\scriptstyle\pm 3.6}$ \\

\quad {\scriptsize $\triangle$ vs. DR-RL}
& {\scriptsize $+6.4\%$}
& {\scriptsize $+14.3\%$}
& {\scriptsize $+8.2\%$} \\ 

\bottomrule
\end{tabular}%
}
\caption{
Performance comparison on Deepresearch benchmarks. Rows prefixed with ``$+$'' denote alternative training strategies initialized from the corresponding SFT checkpoint. D(Warm)L-RL includes an additional long-context RL stage after D(Warm)-RL. 
Best results are bolded, and second-best results are underlined within model families. 
}
\label{tab:deepsearch_comparison}
\end{table}

\subsection{Performance on Deepresearch Benchmarks} 
Tab.~\ref{tab:deepsearch_comparison} compares our models with representative Deepresearch (DR) systems. We make three main observations.
\textbf{(i)} The 4B and 9B models trained with DLD-RL achieve leading performance among models of comparable scale, showing that the framework generalizes across model sizes and backbones.
\textbf{(ii)} DLD-RL consistently outperforms the corresponding DR-RL-only baselines on both backbones and all three benchmarks. This suggests that the intermediate long-context training stage helps overcome the performance plateau of continued standard DR-RL.
\textbf{(iii)} After the intermediate \textbf{D(Warm)L-RL} stage, BrowseComp performance remains similar or slightly decreases, while the other two benchmarks generally improve. This indicates that stronger long-context understanding can already benefit Deepresearch before the final DR-RL stage, particularly on tasks requiring extensive evidence aggregation and cross-document reasoning. In contrast, BrowseComp may depend more on precise search decisions and early exploration. 
Overall, the results support our hypothesis that long-context understanding is an important foundation for further improving DR agents.

\begin{table}[h]
\centering

\resizebox{\columnwidth}{!}{%
  \begin{tabular}{@{}lccc@{}}
  \toprule
  \textbf{Model}
  & \shortstack{\textbf{LongBench}\\\textbf{-v2-128K}}
  & \shortstack{\textbf{Frames}\\\textbf{-128K}}
  & \shortstack{\textbf{LongReason}\\\textbf{-128K}} \\
  \midrule

  Qwen3-30B-A3B-Thinking-2507
  & 47.3 & 76.1 & 79.1 \\
  \midrule
  
\multicolumn{4}{c}{\textit{Qwen3-4B-Thinking-2507 Family}} \\
\addlinespace[2pt]
  Qwen3-4B-Thinking-2507
  & 37.8 & 68.1 & \textbf{68.4} \\

  --\quad AgentCPM-Explore-4B
  & 17.5 & 51.8 & 38.0 \\

  --\quad DR-Venus-4B-SFT
  & 26.4 & 58.6 & 60.3 \\

  DR-Venus-4B-SFT
  & $26.4\,{\scriptstyle\pm 2.3}$
  & $58.6\,{\scriptstyle\pm 3.1}$
  & $60.3\,{\scriptstyle\pm 2.7}$ \\

  \quad $+$ DR-RL Only
  & $28.4\,{\scriptstyle\pm 3.4}$
  & $61.5\,{\scriptstyle\pm 2.2}$
  & $63.1\,{\scriptstyle\pm 3.6}$ \\

  \quad $+$ \textbf{(Ours) D(Warm)-RL}
  & $28.4\,{\scriptstyle\pm 2.8}$
  & $59.2\,{\scriptstyle\pm 3.5}$
  & $60.8\,{\scriptstyle\pm 2.1}$ \\

  \quad $+$ \textbf{(Ours) D(Warm)L-RL}
  & $\mathbf{40.6}\,{\scriptstyle\pm 3.7}$
  & $\underline{68.2}\,{\scriptstyle\pm 2.6}$
  & $\underline{65.3}\,{\scriptstyle\pm 3.2}$ \\

  \quad $+$ \textbf{(Ours) DLD-RL}
  & $\underline{39.8}\,{\scriptstyle\pm 3.1}$
  & $\mathbf{69.9}\,{\scriptstyle\pm 3.8}$
  & $64.6\,{\scriptstyle\pm 2.4}$ \\

  \quad {\scriptsize $\triangle$ vs. DR-RL}
  & {\scriptsize $+40.1\%$}
  & {\scriptsize $+13.7\%$}
  & {\scriptsize $+2.4\%$} \\

  \midrule
  \multicolumn{4}{c}{\textit{Qwen3.5-9B-Instruct Family}} \\
  \addlinespace[2pt]

  Qwen3.5-9B-Instruct
  & $45.5\,{\scriptstyle\pm 2.5}$
  & $75.2\,{\scriptstyle\pm 3.3}$
  & $72.9\,{\scriptstyle\pm 2.8}$ \\

  Qwen3.5-9B-Instruct-DR-SFT
  & $42.1\,{\scriptstyle\pm 3.6}$
  & $73.4\,{\scriptstyle\pm 2.1}$
  & $69.3\,{\scriptstyle\pm 3.4}$ \\

  \quad $+$ DR-RL Only
  & $42.0\,{\scriptstyle\pm 2.7}$
  & $75.1\,{\scriptstyle\pm 3.7}$
  & $71.4\,{\scriptstyle\pm 2.5}$ \\

  \quad $+$ \textbf{(Ours) D(Warm)-RL}
  & $42.8\,{\scriptstyle\pm 3.2}$
  & $74.0\,{\scriptstyle\pm 2.9}$
  & $70.1\,{\scriptstyle\pm 3.8}$ \\

  \quad $+$ \textbf{(Ours) D(Warm)L-RL}
  & $\mathbf{47.1}\,{\scriptstyle\pm 3.9}$
  & $\underline{78.3}\,{\scriptstyle\pm 2.4}$
  & $\underline{77.0}\,{\scriptstyle\pm 3.1}$ \\

  \quad $+$ \textbf{(Ours) DLD-RL}
  & $\underline{46.3}\,{\scriptstyle\pm 2.6}$
  & $\mathbf{78.9}\,{\scriptstyle\pm 3.6}$
  & $\mathbf{78.1}\,{\scriptstyle\pm 2.3}$ \\

\quad {\scriptsize $\triangle$ vs. DR-RL}
& {\scriptsize $+10.2\%$}
& {\scriptsize $+5.1\%$}
& {\scriptsize $+9.4\%$} \\

\bottomrule
\end{tabular}%
}
\caption{
Long-context performance of Deepresearch models and larger reference models.
Rows prefixed by ``$+$'' denote alternative RL training modes initialized
from the corresponding SFT checkpoint. 
D(Warm)L-RL includes an additional long-context RL stage after D(Warm)-RL. 
Within each model family, the best result is shown in bold and the second-best is underlined.}
\label{tab:LongContext_comparison}
\end{table} 
\subsection{Performance on LongContext Benchmarks}
Tab.~\ref{tab:LongContext_comparison} compares the long-context capabilities of our models with representative DR models and their base models. We make three main observations.
\textbf{(i)} DLD-RL consistently improves performance across all three long-context benchmarks on both model scales, demonstrating its effectiveness in strengthening long-input understanding and cross-document evidence integration.
\textbf{(ii)} Existing DR models generally underperform their original base models, suggesting that conventional DR post-training may degrade general long-context capabilities. Standard DR-RL only partially recovers this loss, whereas DLD-RL enables the 9B model to surpass its base model and achieve performance comparable to a much larger 30B-A3B model.
\textbf{(iii)} The final DR-RL stage preserves, and sometimes further improves, the gains obtained from LongQA-RL. Together with the Deepresearch results, this suggests that long-context reasoning and agentic search can reinforce each other. By reusing the model's own interaction trajectories, DLD-RL improves both capabilities through a self-boosting training process.

\begin{table}[t]
\centering
\label{tab:ablation}

\resizebox{\columnwidth}{!}{%
\begin{tabular}{@{}lcc@{}}
\toprule
\textbf{Model / Training Strategy}
& \textbf{Deepresearch Avg.}
& \textbf{Long Avg.} \\
\midrule

\multicolumn{3}{@{}l}{DR-Venus-4B-SFT} \\

\quad $+$ DR-RL Only
& 33.8 & 51.0 \\

\midrule

\quad $+$ \textbf{(Ours) DLD-RL}
& \textbf{35.4} & 58.1 \\

\quad $+$ DLD-RL w/o KL Loss
& \underline{34.6} & \underline{59.4} \\

\quad $+$ DLD-RL w/o Incorrect-Trace QAs
& 34.4 & 57.0 \\

\quad $+$ DR-RL $\rightarrow$ LongQA-RL
& 34.0 & \textbf{59.8} \\

\midrule

\multicolumn{3}{@{}l}{Qwen3.5-9B-Instruct-DR-SFT} \\

\quad $+$ DR-RL Only
& 37.9 & 62.8 \\

\midrule

\quad $+$ \textbf{(Ours) DLD-RL}
& \textbf{41.5} & 67.8 \\

\quad $+$ DLD-RL w/o KL Loss
& \underline{40.6} & \underline{69.0} \\

\quad $+$ DLD-RL w/o Incorrect-Trace QAs
& 40.2 & 66.8 \\

\quad $+$ DR-RL $\rightarrow$ LongQA-RL
& 38.9 & \textbf{69.4} \\

\bottomrule
\end{tabular}%
}
\caption{
Ablation study of DLD-RL on Deepresearch and long-context benchmarks.
Rows prefixed with ``$+$'' denote alternative training strategies initialized
from the corresponding SFT checkpoint.
Results are averaged over the three benchmarks in each category. 
}
\end{table}

\subsection{Ablation Study}
We conduct three ablation studies by comparing our full method with the original model, the DR-RL-only baseline, and the intermediate checkpoints corresponding to different training stages. Specifically, we examine the following variants: \textbf{(i)} removing the KL regularization used during LongQA-RL; \textbf{(ii)} excluding long-context QA instances derived from failed rollout trajectories; and \textbf{(iii)} moving LongQA-RL to after the full Deepresearch RL stage. 

For \textbf{(i)}, the multi-turn agentic behavior acquired through SFT is inherently fragile. Removing KL regularization allows the policy to deviate more aggressively toward the LongQA objective, which can produce larger gains in long-context capability but simultaneously degrades the agentic behaviors required for Deepresearch. This result highlights the importance of constraining policy drift during the intermediate LongQA-RL stage.
For \textbf{(ii)}, failed rollout trajectories typically contain more irrelevant, misleading, or mutually conflicting documents than successful trajectories. In contrast, contexts constructed solely from successful rollouts often contain relatively straightforward evidence structures and therefore provide insufficient difficulty for training robust long-context reasoning. Incorporating failed trajectories exposes the model to harder distractors and more complex cross-document relationships, leading to stronger training signals.
For \textbf{(iii)}, our ultimate goal is to improve Deepresearch performance. Once DR-RL has reached a performance plateau, shifting the optimization objective to LongQA may partially weaken the agentic capabilities acquired during the preceding DR-RL stage. The final Deepresearch RL stage therefore serves as an annealing phase, gradually realigning the policy with the Deepresearch objective and translating the enhanced long-context capability into more effective search and browsing behavior. 
Notably, across all insertion positions and data variants, DR-to-Long data consistently improves long-context QA performance, demonstrating the effectiveness of the constructed training data. 

\subsection{The Potential of Rollouts to LongContext-QAs}
\begin{figure}[h]
    \centering
    \resizebox{\columnwidth}{!}{
        \includegraphics{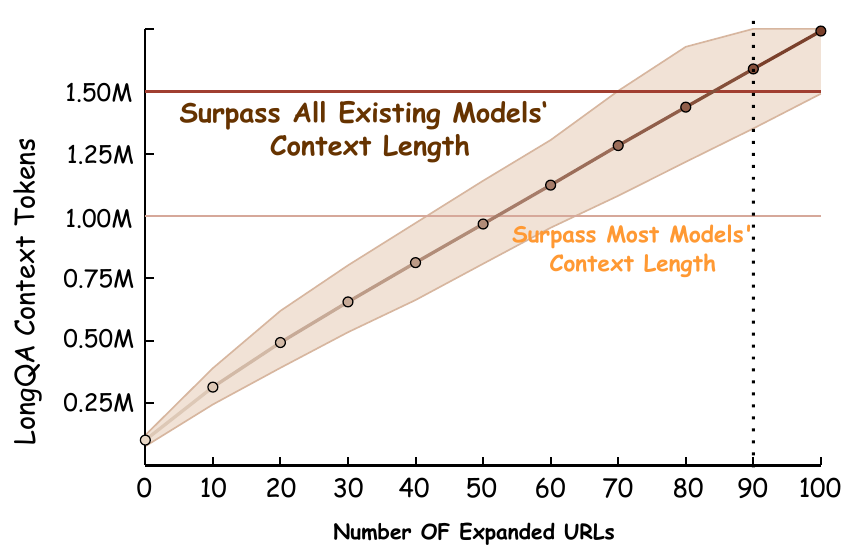}
    }
    \caption{Growth of the constructed LongQA context length as more URLs from DeepResearch rollouts are expanded.}
    \label{fig:context_potential}
\end{figure}

The core idea of DR Rollouts to LongContext QA (DR-to-Long) is to replace compact rollout observations, such as search snippets and webpage summaries, with the full contents of their corresponding URLs. This preserves the original semantic and evidential relationships among retrieved documents while substantially increasing the context length. In our main experiments, we truncate the constructed contexts to 64K--128K tokens to match the training configuration. In principle, however, their length is limited only by the total content of the visited webpages. 
To evaluate the scalability of this construction, we progressively expand more URLs from real DR rollouts and measure the resulting context length on Fig.~\ref{fig:context_potential}. 
Context length grows rapidly with expanded URLs, reaching about 1.5 million tokens for 100 URLs—beyond most public models’ limits. Since such data can be generated from existing rollouts with minimal annotation, DR trajectories offer a natural source for long-context pretraining.

\subsection{LongContext Ability Influnces the Deepresearch Tool-use Preference}
We further investigate how LongQA-RL affects the agent’s search behavior by comparing its tool-use statistics with those of the DR-RL-only baseline. After LongQA-RL, the model generates fewer redundant queries in \textsc{Search} and uses the \textsc{Visit} tool more frequently. We consider two search queries similar when the Jaccard similarity between their terms is at least 0.8. This behavioral shift suggests that the agent relies less on repeatedly exploring similar information and instead conducts more in-depth examination of promising sources. It also reflects a better understanding of the accumulated context, enabling the agent to avoid revisiting previously explored evidence and allocate its tool budget more effectively. Overall, these results indicate that stronger long-context ability can lead to more focused and efficient evidence-gathering behavior. 
\begin{table}[h]
\centering
\resizebox{\columnwidth}{!}{%
\begin{tabular}{@{}lccc@{}}
\toprule
\textbf{Model}
& \textbf{Avg. Search}
& \textbf{Avg. Visit}
& \shortstack{\textbf{Similar Search} \textbf{(\%)}} \\

\midrule

DR-Venus-4B-SFT
& -- & -- & -- \\

\quad $+$ DR-RL Only
& 56.62 & 12.72 & 27.56 \\

\quad $+$ DLD-RL
& 57.22 & 14.39 & 25.22 \\

\midrule

Qwen3.5-9B-Instruct-DR-SFT
& -- & -- & -- \\

\quad $+$ DR-RL Only
& 47.96 & 11.26 & 23.69 \\

\quad $+$ DLD-RL
& 49.57 & 21.34 & 18.14 \\

\bottomrule
\end{tabular}%
}

\caption{Tool-use behaviors on BrowseComp. Rows prefixed by ``$+$'' denote successive training stages initialized from the model immediately above.}
\label{tab:tool_behavior}
\end{table}

\section{Conclusion}

In this work, we identify insufficient LongContext capability as a key bottleneck for DeepResearch agents. We propose DR-to-Long and DLD-RL. DR-to-Long converts naturally collected DeepResearch rollout data into multi-document LongContext QA data without additional annotation, and also has the potential to provide ultra-long-context data for pretraining. DLD-RL is a DR $\rightarrow$ LongQA $\rightarrow$ DR training framework. Experiments show that DLD-RL consistently improves both DeepResearch performance and LongContext reasoning. Further analysis shows that stronger LongContext capability changes agents’ tool-use preferences, suggesting an inherent connection between the two capabilities. 
\bibliography{aaai2027}


\end{document}